# UNSUPERVISED DEEP TRANSFER FEATURE LEARNING FOR MEDICAL IMAGE CLASSIFICATION

*Euijoon Ahn[1], Ashnil Kumar[1], Dagan Feng[1,4], Michael Fulham[1,2,3], and Jinman Kim[1]*

[1]School of Computer Science, University of Sydney, Australia
[2]Department of Molecular Imaging, Royal Prince Alfred Hospital, Australia
[3]Sydney Medical School, University of Sydney, Australia
[4]Med-X Research Institute, Shanghai Jiao Tong University, China

## ABSTRACT

The accuracy and robustness of image classification with supervised deep learning are dependent on the availability of large-scale, annotated training data. However, there is a paucity of annotated data available due to the complexity of manual annotation. To overcome this problem, a popular approach is to use transferable knowledge across different domains by: 1) using a generic feature extractor that has been pre-trained on large-scale general images (i.e., transfer-learned) but which not suited to capture characteristics from medical images; or 2) fine-tuning generic knowledge with a relatively smaller number of annotated images. Our aim is to reduce the reliance on annotated training data by using a new hierarchical unsupervised feature extractor with a convolutional auto-encoder placed atop of a pre-trained convolutional neural network. Our approach constrains the rich and generic image features from the pre-trained domain to a sophisticated representation of the local image characteristics from the unannotated medical image domain. Our approach has a higher classification accuracy than transfer-learned approaches and is competitive with state-of-the-art supervised fine-tuned methods.

## 1. INTRODUCTION

The rapidly increasing collections of diverse medical images is a direct consequence of the central role that medical imaging plays in modern healthcare. These massive image collections are a valuable source of knowledge, provide the opportunity for evidence-based clinical decision-making, clinician training, biomedical research [1], and enable computer-aided diagnosis systems (CADs) for the automated identification, retrieval and classification of imaging data [2]. These CADs associate low-level image features to high-level expert domain knowledge (e.g., annotations, disease stage, etc.) using supervised machine learning techniques [3].

While many different types of medical images are collected [4, 5], the source of the images are not always identified appropriately [6, 7]. Automatic identification of the imaging modality is an initial fundamental requirement as semantics and content of an image can vary greatly depending on the modality. The routine automated classification of the imaging modality, however, is not trivial due to subtle variations in imaging types (e.g., different types of anatomical images), and appearance of images based on the individual disease that is depicted [3].

With the availability of large-scale annotated image dataset (e.g., ImageNet), supervised deep learning approaches such as convolutional neural networks (CNN) have enabled highly representative image features to be derived for numerous computer vision researches [8, 9]. CNNs learn image features in a hierarchical manner, generally capturing more sophisticated image representations with deeper networks [8]. Such supervised approaches, however, are problematic in the medical domain where large-scale annotated datasets are very rare. Medical images are complex to interpret, very time-consuming to annotate and are also very dependent of the overall expertise of the medical imaging.

The learning transferable knowledge across different domains is one approach to address the lack of available annotated medical image data. A generic feature extractor that has been pre-trained on a large-scale general images (i.e. transfer-learned) and fine-tuning the generic knowledge towards tasks with considerably small set of annotated medical images have been employed [3, 10, 11]. Transfer-learned features rely upon more general features, so they are not able to capture domain-specific characteristics of medical images. Any fine-tuning approach to address these limitations is still dependent upon the availability of sufficient annotated data. Unsupervised feature learning algorithms that learn image features from unlabelled data are one way to address these limitations. The features are usually learned using sparse coding [12] and auto-encoder (AE) algorithms [13] but both have often been limited to learning only low-level local structures such as lines and edges. This is mainly attributed to using simple decomposed image patches that cannot effectively learn the global structures and local connections between image content [14, 15]. Convolutional sparse coding (CSC) and convolutional auto-encoders (CAEs) extend the original patch-based models to cope with multidimensional and large-sized images. Both have performed well in natural image reconstruction, denoising, and classification [16, 17]. CAEs, in particular, can learn global structures, using multidimensional filters with convolutional operation, and unlike patch-based methods, they preserve the relationships between neighbourhood and spatial information.

In this paper, we propose a new hierarchical unsupervised feature extractor by introducing a convolutional auto-encoder placed atop of a pre-trained CNN. The CAE layer transforms the feature maps from the pre-trained network to a set of non-redundant and relevant medical image features. Hence, our architecture can constrain the rich generic image features from a pre-trained domain, for example annotated natural images from ImageNet, to a sophisticated representation of the local image characteristics of the unannotated medical image domain. We validated our approach on a public dataset and compared it to other unsupervised approaches and state-of-the-art supervised methods.

## 2. METHODS

### 2.1. Overview of Unsupervised Deep Transfer Feature Learning

Our architecture is shown in Fig. 1. A convolutional layer is added on top of a pre-trained CNN from a different domain, where the weights of the layer are learned using the CAE. This approach preserves generic image features and captures specific local characteristics that lie within the medical images. As in a standard CNN, a rectified linear unit (ReLU) layer is added to have non-negative feature maps. The final feature representation is then extracted in a feed-forward manner.

### 2.2. Convolutional Auto-Encoders

An AE has an encoder and a decoder that take an input $x \in \mathcal{R}^m$ and processes it to the latent representation of using a deterministic function $\mathbf{Z} = \sigma(W_e x + b_e)$, where $\theta_e = \{W_e, b_e\}$ are the parameters of the encoder. The code $z$ is then reversed mapped using a function $\mathbf{y} = \sigma(W_d z + b_d)$, where $\theta_d = \{W_d, b_d\}$ are the parameters of the decoder and $\sigma$ is the non-linear activation function. The two parameter sets are generally constrained to have a form of $W_d = W_e^T$, preventing to learn degenerated features. For each input $x_i$ is then mapped back to its code $z_i$ and its reconstruction $y_i$. The parameters are usually optimized by minimising a loss function over the training data set.

Classic AEs ignore the global structures and local connections between image content, which produce redundancy in the parameters. Unfortunately this makes it difficult for each feature to capture sophisticated localised information within training image dataset. CAEs, on the other hand, capture and share localised information among all locations in the input.

The CAE architecture is similar to classic AEs, except that the weights are shared. For an input $x \in \mathcal{R}^{m \times n}$, the output of $k$th feature map can be expressed as:

$$z^k = \sigma(x * W_e^k + b_e^k) \quad (1)$$

where $\sigma$ is ReLU activation units and $*$ denotes 2D convolution. Here, the bias is processed per feature map. The reconstruction is then conducted as follows [18]:

$$\mathbf{y} = \sigma\left(\sum_{k \in M} z^k * \widetilde{W}_d^k + b_d^k\right) \quad (2)$$

where $\widetilde{W}$ is 180º flipped weight matrix and $M$ represents the group of latent feature maps. The cost function to minimise the loss is implemented as:

$$\mathrm{E}(\theta) = \min_{W\ b} \frac{1}{2} \sum_{i=1}^{d} \|x_i - y_i\|_2^2 \quad (3)$$

As in a standard backpropagation algorithm, the reconstruction error gradient is first back propagated and then the weights are updated using stochastic gradient descent.

### 2.3. Zero-bias Activation at Encoding-time

Training a CAE often generates hidden unit biases that take large negative values. The negative values are a natural result of using a hidden bias layer that represents the input data and controls the sparsity of the representation. These hidden unit biases, however, have a detrimental effect as they make it difficult to learn non-trivial image characteristics [19, 20]. When considering ReLU units and negative biases in an AE, the model will tend to learn a point attractor to represent image features from a restricted space rather than multidimensional regions. Thus, we use a zero-bias ReLU activation that fixes the biases ($b$) of our convolutional and deconvolutional layers to zero at encoding-time.

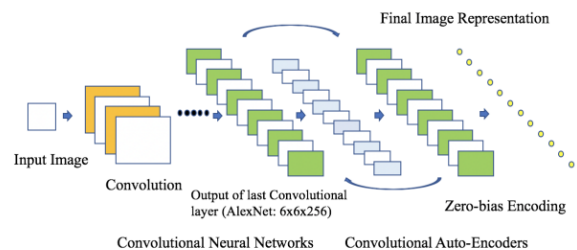

**Fig. 1.** An illustration of our architecture.

## 2.4. Domain Adaptation: Integration of a Pre-trained CNN

We used the well-established pre-trained CNN AlexNet [21] to extract rich generic features from medical images. AlexNet takes as inputs, RGB images with a size of 227 x 227. We therefore re-scaled our medical images to this resolution and colour space (only if images are not RGB). We used the output of the last convolutional layer of the pre-trained CNN as an input to subsequent CAE learning (the input size is therefore 6 x 6 x 256).

## 3. MATERIALS AND EXPERIMENTAL SETUP

We used the medical Subfigure Classification dataset used in the ImageCLEF 2016 competition [7]. The dataset has 6776 training images, 4166 test images from 30 different image modalities, and ground annotations are available for both image datasets. As the baseline, we compared our approach to transfer-learned approaches based on the pre-trained AlexNet [21] and GoogLeNet [22]. For this analysis, we used the results reported by Kumar et al. [3]. We also compared our approach to the best supervised approaches based on an ensemble of fine-tuned models [3, 10]. We also made further comparisons with the following established unsupervised feature learning approaches: sparse coding (SC); Independent Component Analysis (ICA); and stacked sparse auto-encoders (SSAE). We used the Top 1 accuracy (the correctness of the predicted label) that is the standard performance measure adopted in recent CNN studies for the classification of medical image modalities [3].

For all the learned features (SC, ICA, SSAE, and our method), we used the multi-class linear SVM with a differentiable quadratic hinge loss, used by Yang et al. [23], so that the training could be done with simple gradient-based optimisation methods. We used LBFGS with a learning rate of 0.1 and a regularization parameter of 1, consistent with the parameters specified by Yang et al. [23].

### 3.1. Implementation Details

Our CAE was trained for 100 epochs with a batch size of 512 and an initial learning rate of $10^{-5}$. We used learning rate annealing, decaying the rate by a factor of 10 when the error plateaued on a GeForce GTX 1080 Ti GPU (11GB memory).

**Table 1.** The network parameters of our CAE.

| Kernel | 3x3 |
|---|---|
| Conv Stride | 1 |
| Pad | 1 |
| Pool | 2 |
| Filter | 4096 |

Table 1 shows the network parameters. The main parameter in our architecture is the size of filters in a CAE network. We used an empirical process to discover appropriate filter size (512, 1024, 2024 and 4096) and set 4096 in our all experiments.

## 4. RESULTS AND DISCUSSION

Our classification accuracy is shown in Table 2; it has greater accuracy than the unsupervised feature learning algorithms, the baseline transfer-learned and the fine-tuned CNN features with a top 1 accuracy of 81.33%. The best performing approach was the ensemble of fine-tuned GoogLeNet and ResNet with an accuracy of 83.14%. So our method had competitive accuracy with state-of-the-art supervised CNNs.

We attribute this performance to: 1) being able to leverage generic image characteristics in the pre-trained CNN and the specific local characteristics of medical images learned via CAEs; 2) the 2D convolution operation and, 3) zero-bias activation that allows effective learning of the global structure and local relationships between image contents. Conventional unsupervised approaches such as SC and SSAE do not consider the relationships between neighbourhood and spatial information and so cannot extract discriminative image features. While transfer-learned CNNs were able to extract more generic and representative image features, they could not build data-specific features. As expected, the fine-tuning of AlexNet and GoogLeNet produced improved results (as in [3]) relative to the baseline transfer-learned methods; however, the performances of these approaches were dependent on the size of annotated dataset. Our approach, on the other hand, was insensitive to these issues and could derive domain-specific image features without reliance on labels, with a performance that was better than fine-tuned AlexNet and GoogLeNet. With the additional manual data expansion (almost a double size of the original dataset), the ensemble of GoogLeNet and ResNet was able to achieve top 1 accuracy of 87.87% [10]. We suggest that our method would also improve given additional training data. Our results show that the larger size of filters improve the final feature representation at the cost of increased computational complexity (see Table 3).

**Table 2.** Top 1 Image classification results on ImageCLEF16 dataset with comparison with conventional unsupervised methods as well as other state-of-the-art supervised approaches.

| Methods | Top 1 Accuracy (%) |
|---|---|
| Sparse Coding | 57.08 |
| ICA | 58.79 |
| Stacked Sparse Auto-encoder (2 layers) | 65.17 |
| Transfer-learned GoogLeNet + SVM [22] | 78.61 |
| Transfer-learned AlexNet + SVM [21] | 79.21 |
| Fine-tuned AlexNet + SVM [3] | 79.60 |
| Fine-tuned GoogLeNet + SVM [3] | 80.75 |
| Our method | 81.33 |
| An ensemble of fine-tuned AlexNet and GoogLeNet [3] | 82.48 |
| An ensemble of fine-tuned GoogLeNet and ResNet [10] | 83.14 |

**Table 3.** Results of classification performance with different filter sizes.

| Filter size | Top 1 Accuracy (%) |
|---|---|
| 512 | 77.00 |
| 1024 | 79.48 |
| 2048 | 80.15 |
| 4096 | 81.33 |

Ensemble approaches [3, 10] that can learn and represent different image characteristics coupled with our framework could potentially extract better image representations. We will explore such an approach in future work.

## 5. CONCLUSION

In this paper we propose a new hierarchical unsupervised feature extractor by introducing a zero-bias CAE placed atop of a pre-trained CNN. Our framework constrains the rich generic image features from the pre-trained domain to a sophisticated representation of the local image characteristics of the unannotated medical image domain. We evaluated our framework by comparing it to other supervised approaches using a public medical dataset. Our results show that our method is competitive with the state-of-the-art supervised CNNs, indicating that it enables improved feature representation of medical imaging data in an unsupervised manner. We suggest that our approach can benefit many medical image analysis tasks when there are none or limited annotated training data.